\documentclass[preprint,12pt]{elsarticle}

\usepackage{amsmath,amssymb,amsfonts}
\usepackage{mathtools}
\usepackage{bm}
\usepackage{siunitx}

\usepackage{booktabs}
\usepackage{array}
\usepackage{tabularx}
\usepackage{multirow}
\usepackage{ragged2e}
\newcolumntype{L}{>{\RaggedRight\arraybackslash}X}
\usepackage{adjustbox}

\usepackage{subcaption}
\usepackage{enumitem}

\usepackage{tikz}
\usetikzlibrary{positioning,arrows.meta,fit,shapes.geometric,calc}

\usepackage[hyphens]{url}

\biboptions{sort&compress}
\journal{Biomedical Signal Processing and Control}

\begin{document}

\begin{frontmatter}

\title{A Montage-Agnostic Encoder for Calibration-Light Cross-User Gesture Recognition from
  Surface Electromyography}

\author[bme]{Jethro Odeyemi}
\ead{jethro.odeyemi@usask.ca}

\author[bme]{W.J. (Chris) Zhang\corref{cor1}}
\ead{chris.zhang@usask.ca}
\cortext[cor1]{Corresponding author.}

\affiliation[bme]{organization={Division of Biomedical Engineering, University of Saskatchewan},
            addressline={57 Campus Drive},
            city={Saskatoon},
            state={SK},
            postcode={S7N 5A9},
            country={Canada}}

\begin{abstract}
Pattern-recognition control promises a myoelectric prosthesis that responds to many intended
gestures rather than one or two, but the promise has stayed in the laboratory. A recogniser
trained on one person rarely transfers to the next, and useful performance usually demands a
fresh round of labelled calibration from the end user. A montage-agnostic encoder is introduced
that reads each electrode with shared weights and locates it by its physical coordinate rather
than its index, so one architecture ingests any channel count without montage-specific
parameters. Trained across users, it exceeds a per-user Hudgins and linear-discriminant
classifier by 0.234 macro-F1 on DB1 for every held-out subject and by 0.108 on DB2, and falls
below it on the ten-subject DB5. Each of the encoder's three key components
individually accounts for more than half of its 3-shot macro F1 in an otherwise budget-matched
ablation study. A controlled subject-count sweep shows the margin is close to flat from nine
training subjects to thirty-nine, so the training pool binds only as a stability floor below
which cross-user training fails to converge; what tracks the direction of the comparison
across the three databases is instead the strength of the per-user baseline, which signal
fidelity sets.
Comparing against an LDA baseline depends on budget spent training models and on how
good that baseline is, and self-supervised pretraining had no benefits once a supervised model
was adequately trained.
\end{abstract}

\begin{keyword}
surface electromyography \sep gesture recognition \sep cross-user transfer \sep calibration
\sep transformer encoder \sep upper-limb prosthetic control
\end{keyword}

\end{frontmatter}

\section{Introduction}
\label{sec:ch3-intro}

The majority of surface-EMG based gesture recognition systems require a calibration phase to fit
a specific decoder to the user's signal. A typical calibration involves the user holding a
sequence of predefined postures while the system learns the user's signal characteristics. The
calibration process is time consuming and requires repeated iterations if the electrodes are
moved. Therefore, reducing calibration costs is among the long standing issues within this
community~\citep{odeyemi_review}.

Because of differences in anatomy and electrode placements, surface-EMG signals vary
significantly among individuals. Hence, deep neural network models trained on one group of users
tend to generalize poorly to other
groups~\citep{coteallard2019deep,coteallard2020interpreting,lobov2018latent,shen2022generalization}.
As a result, researchers have largely relied on per-user calibration to bridge the gap. This
paper aims to reduce the amount of calibration required by incorporating sufficiently powerful
prior knowledge about users into a model. If successful, we expect to reduce the number of
calibration repetitions from a full screen-guided session down to a handful.

There have been two major challenges to adopting general-purpose encoders for this task. First,
electrode configurations differ widely between devices/studies. Some studies have used 8-channel
arm bands, others have employed 12-channel arm bands or a 16-channel band. Given the different
geometric structures in each case, the input layer of any model that assumes a fixed channel
count cannot be shared across them. Second, calibration itself remains a challenging issue: the
default per-user linear classifiers on hand-crafted features perform reasonably well at small
amounts of data and provide a difficult-to-surpass baseline.

In this paper, we present an encoder design intended to address both of the aforementioned
issues and evaluate it using a protocol that measures cross-user recognition performance versus
calibration budget. Specifically, our goal is to measure how well a model recognizes gestures
performed by users that were not included in its training set, as a function of how many
repetitions the unknown user provides for calibration. Our proposed encoder views each electrode
as a token located at its corresponding position on the forearm. We also mix together multiple
electrodes via an attention mechanism, thereby allowing the model to learn the spatial patterns
of activity rather than focusing solely on any single channel. Furthermore, we normalize each
channel independently via causal running means and variances, enabling the model to adapt to any
new user's EMG signals at run time with no additional labeled data needed. Finally, we also
include a channel-masking regularization term to encourage the model to focus on multi-electrode
features that are less sensitive to channel failures/movements. Thus, our main contributions are
as follows:

\begin{enumerate}
  \item \textit{Montage-Agnostic Encoder} (Section~\ref{sec:ch3-method}): An encoder
    architecture capable of decoding gestures from surface EMG signals collected from four
    different electrode arrays (eight to sixteen channels) with no montage-specific parameters.
    We train a separate model on each dataset and only train a single encoder jointly across
    montages in the self-supervised study presented in Section \ref{sec:ch3-pretrain}. An
    ablation analysis is conducted to assess the relative contributions of each of the three key
    components of the encoder architecture.
  \item \textit{Evaluation of Calibration Efficiency} (Section~\ref{sec:ch3-results-main}):
    Results demonstrating that, given an adequate training budget, our encoder outperforms the
    per-user LDA classifier on the entire NinaPro DB1 benchmark for both calibration budgets
    where the latter can be trained (one and three repetitions), and on all twenty-seven
    held-out users.
  \item \textit{Characterization of When Learned Encoders Outperform Per-User Classifier
    Baseline vs. When They Do Not}: Section \ref{sec:ch3-interaction} examines under what
    circumstances our learned encoder outperformed the per-user LDA classifier and when it
    failed to do so. That section demonstrates that the operative factor is the quality of the
    signal provided to the LDA classifier, specifically signal fidelity, which determines how
    well that classifier performs. Separately, we conduct a controlled subject-count sweep that
    demonstrates that the size of the training pool serves only as a stability floor for our
    learned encoder. This leads us to make a prediction regarding future work on using learned
    encoders in amputee settings (tested in companion work \citep{odeyemi_amputee}), which we
    partially verify.
  \item Section \ref{sec:ch3-pretrain} discusses two self-supervised pre-training objectives
    that did not lead to improved results for our learned encoder. That section describes why
    those objectives failed.
\end{enumerate}

\section{Background and positioning}
\label{sec:ch3-background}

Surface-EMG gesture recognition research using deep learning techniques progressed rapidly from
2015--2020 on top of the NinaPro database
collections~\citep{coteallard2019deep,atzori2016deep,hu2018novel,zhai2017self}. Window-based
CNNs~\citep{atzori2016deep} and instantaneous EMG image CNNs~\citep{geng2016gesture} established
themselves as state-of-the-art benchmarks for within-subject accuracy, and transfer-learning
versions of those architectures diminished, but did not eliminate, the necessity for per-user
calibration~\citep{coteallard2019deep}. Prior to those works and continuing today in clinical
environments sit linear discriminant classifiers based upon the Hudgins feature
set~\citep{hudgins1993new}, i.e., mean absolute value, waveform length, zero crossings, and
slope sign changes. That pipeline is considered our comparison baseline throughout this paper
since clinics continue to deploy
it~\citep{phinyomark2018feature,wu2016wearable,hargrove2017myoelectric,li2017motion} and since
it operates effectively at small calibration budgets.

Recent research has sought to advance cross-user recognition using large pretrained encoders
along with flexible tokenization for surface EMG electrode configurations; some of that research
has utilized architectures that closely resemble ours. For example, \citet{saft2025}'s Spatially
Aware Feature-learning Transformer shares this paper's central architectural philosophy:
per-channel temporal feature extraction, learned encoding of electrode coordinates, and
cross-channel attention mechanisms. Also related to our approach are split-encoder strategies
for surface EMG processing~\citep{tinymyo2025,hadidi2025splashnet}, causal input normalization
strategies, and channel-masked regularization strategies. While our encoder was designed for
calibration-light cross-user scenarios and optimized accordingly, Section~\ref{sec:ch3-ablation}
conducts an ablation analysis to quantify each component's effectiveness. This paper claims no
new architecture class; it demonstrates that, given adequate training resources, this particular
architecture achieves better performance than per-user LDA baselines at all calibration budgets
where such a baseline can be trained (one and three repetitions); \citet{saft2025} demonstrated
cross-user balanced accuracy values of 81.8\% compared to 82.9\% achieved by the per-user
classifier; further described in Section~\ref{sec:ch3-interaction} are conditions under which
any encoder will either achieve better or worse results than those obtained by a per-user
classifier.

\section{Method}
\label{sec:ch3-method}

\subsection{Montage-agnostic encoder architecture}

\begin{figure}[tp]\centering
\begin{subfigure}{\textwidth}\centering
\resizebox{\textwidth}{!}{%
\begin{tikzpicture}[
  font=\small,
  box/.style={draw, rounded corners, align=center, minimum height=8mm, inner sep=4pt, fill=black!4},
  op/.style={draw, rounded corners, align=center, minimum height=8mm, inner sep=4pt, fill=blue!7},
  hl/.style={draw, rounded corners, align=center, minimum height=8mm, inner sep=4pt, fill=green!8},
  ->, >=Latex, node distance=6mm and 9mm]

\node[box] (in) {Multichannel sEMG\\window $x\in\mathbb{R}^{C\times T}$\\($C\in\{8,10,12,16\}$)};
\node[op, right=of in] (rtn) {Rolling-time\\normalisation\\(causal, per channel)};
\node[op, right=of rtn] (cnn) {Shared per-channel\\temporal CNN\\(weights tied across $C$)};
\node[hl, right=of cnn] (pos) {$+$ electrode-coordinate\\position encoding\\$\mathrm{MLP}(x_c,y_c,z_c)$};

\node[op, below=17mm of pos] (acm) {Channel\\masking (train only,\\bounded, keep $\ge 3$)};
\node[op, left=of acm] (tf) {Cross-channel\\Transformer\\($N$ layers, attention\\over $C$ tokens)};
\node[op, left=of tf] (pool) {Attention\\pooling over\\channel tokens};
\node[box, left=of pool] (out) {Window\\embedding\\$z\in\mathbb{R}^{d}$};

\draw (in) -- (rtn);
\draw (rtn) -- (cnn);
\draw (cnn) -- (pos);
\draw (pos.south) -- (acm.north);
\draw (acm) -- (tf);
\draw (tf) -- (pool);
\draw (pool) -- (out);

\node[align=center, font=\footnotesize, above=3mm of cnn] {\emph{one token per electrode}};
\node[align=center, font=\footnotesize, fill=white, inner sep=1pt, below=3mm of tf]
  {\emph{position from geometry, not channel index}};

\node[hl, below=17mm of out] (head) {Linear gesture head};
\node[hl, right=7mm of head] (mae) {Envelope-reconstruction\\head (pretraining)};
\draw (out.south) -- (head.north);
\draw[dashed] (out.south) -| (mae.north);
\end{tikzpicture}
}
\caption{}\label{fig:ch3-design-encoder}
\end{subfigure}

\vspace{2ex}

\begin{subfigure}{\textwidth}\centering
\resizebox{0.80\textwidth}{!}{%
\begin{tikzpicture}[
  font=\small,
  subj/.style={draw, rounded corners, minimum width=8mm, minimum height=7mm, inner sep=2pt, align=center},
  train/.style={subj, fill=blue!12},
  test/.style={subj, fill=red!14},
  cal/.style={subj, fill=orange!22, align=center, minimum width=24mm},
  lbl/.style={font=\footnotesize, align=center, fill=white, inner sep=1.5pt},
  ->, >=Latex]

\foreach \i in {1,...,8} { \node[train] (s\i) at (\i*0.95,0) {$s_{\i}$}; }
\node[train] (sd) at (9*0.95,0) {$\cdots$};
\node[test]  (sk) at (10*0.95,0) {$s_k$};
\node[lbl, left=3mm of s1, align=right] {subjects};
\node[lbl] at (5.2,0.85) {base training pool (all subjects except $s_k$)};
\node[lbl] at (5.2,-0.9) {montage-agnostic encoder $+$ gesture head};

\node[test] (hk) at (5.2,-2.3) {$s_k$};
\draw[->] (sk.south) .. controls +(0,-1.1) and +(2.6,0) .. node[lbl, above right, pos=0.55] {held out} (hk.east);
\node[cal]  (c1) at (3.3,-3.9) {calibration\\reps $1,3,4$};
\node[test, minimum width=24mm] (t1) at (7.1,-3.9) {test reps\\$2,5,7$};
\draw[->] (hk.south) -| (c1.north);
\draw[->] (hk.south) -| (t1.north);

\node[lbl] at (5.2,-5.2)
  {budgets: \emph{0-shot} $\to$ \emph{1-shot} $\to$ \emph{3-shot};
   adapt on $k$ calibration reps, score on the disjoint test reps};
\end{tikzpicture}
}
\caption{}\label{fig:ch3-design-protocol}
\end{subfigure}
\caption[Encoder and evaluation protocol]{Encoder and evaluation protocol. (a) The proposed montage-agnostic encoder. Each electrode
becomes a single token whose position is supplied by its forearm coordinate rather than its channel
index, so one set of weights ingests any electrode count. Rolling-time normalisation makes the
representation calibration-free, cross-channel attention mixes the electrode tokens, and bounded
channel masking regularises toward user-invariant features. The linear head is used for gesture
classification; the dashed head is the self-supervised pretraining branch.
(b) Cross-user evaluation. The encoder is trained on every subject except
the held-out one; that subject is then split by repetition into calibration reps and test reps. At
each calibration budget ($0$, $1$, $3$ repetitions) the model is adapted on the calibration reps and
scored on the disjoint test reps, with per-repetition majority voting. No held-out-subject data
reaches base training or model selection.}
\label{fig:ch3-design}
\end{figure}
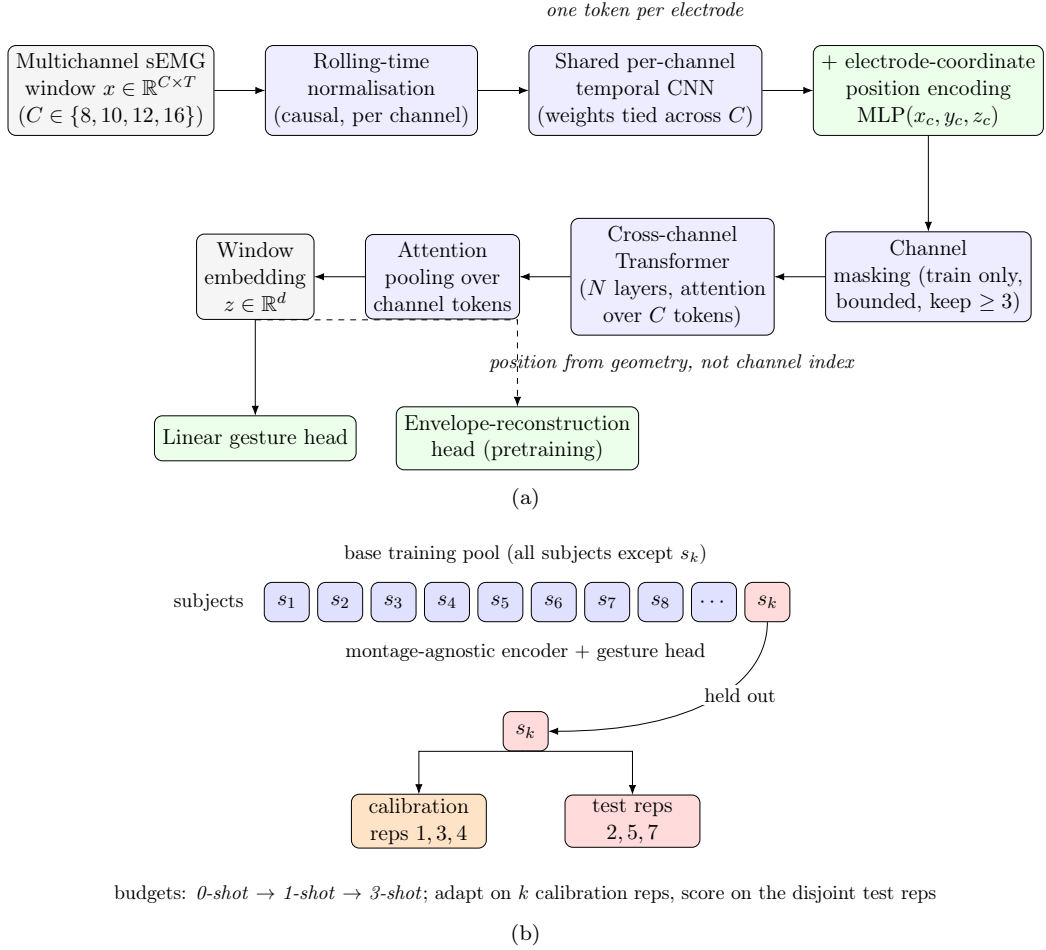

We begin by providing details concerning our encoder's design. The overall architecture is
depicted in Figure~\ref{fig:ch3-design}(a). Our encoder maps sequences of EMG signals ($x \in
\mathbb{R}^{C\times T}$) onto fixed-size embeddings ($z$), where $C$ represents the possible
number of channels in any of the available montages.

Our proposed encoder is founded on the idea that each electrode is treated as one independent
token whose identity corresponds to its physical position on the forearm.

\paragraph{Causal Rolling-Time normalisation.} Our first layer normalizes each channel
individually using their respective causal rolling statistics over time; these statistics
represent expanding window means and variances calculated over time. Since our normalizing is
done per channel and only uses causal information from past time steps, at runtime a previously
unobserved user will self-normalize their input data with respect to their own signal using only
their own causal history; no user-specific statistics will be transported from training to
testing. Hence, even though there is no labeled data from the new user available at runtime, our
model will still adaptively scale its input representations according to new users' signals
without requiring any additional labels.

\paragraph{Shared per-channel tokeniser.} To produce one feature vector per electrode, we apply
a single one-dimensional convolutional stack separately to every channel. Due to sharing of
weights across channels, adding or removing an electrode simply adds/removes a token to/from our
tokenizer without modifying any parameter(s).

\paragraph{Electrode-coordinate position encoding.} Each electrode contains normalized $(x,y,z)$
coordinates specifying its position on the forearm. These coordinates are mapped to vectors
added to each electrode token by means of a multilayer perceptron. Therefore, our model learns
where each electrode resides, not merely its index, thus allowing any electrode configuration to
map into one uniform representation space, with each electrode serving as an unordered element
within that space.

\paragraph{Cross-channel attention.} Our transformer encoder~\citep{vaswani2017attention}
applies attention mechanisms across our electrode tokens. The temporal structure is encoded via
our per-channel tokenizers; attention mechanisms capture spatial relations between electrodes.
Following application of attention mechanisms across all channels, an attention-pooling layer is
used to collapse the electrode tokens into a single window embedding; finally a linear head
computes gesture logits.

\paragraph{Channel-masking regularisation.} During training time, a bounded number of electrodes
is randomly masked such that at least three channels remain active for each montage. This forces
our model towards features that are invariant or robust with regards to missing or displaced
electrodes rather than relying on any single channel. How many electrodes should be masked
affects whether masking fails completely: if too aggressive on very small montages (e.g., all
electrodes masked), there are no valid windows containing active electrodes resulting in
collapsed training; we solved this issue by enforcing bounds for masking.

\subsection{Calibration-light evaluation protocol}
\label{sec:ch3-protocol}

All evaluations are conducted cross-user (leave-one-subject-out), as illustrated in Figure
\ref{fig:ch3-design}(b). Training occurs exclusively on all subjects excluding one; this held-out
subject is divided into calibration repetitions and test repetitions. The model is
updated/calibrated on the calibration repetitions and evaluated on disjoint test repetitions. At
each calibration budget (zero, one, or three repetitions), predictions are made using majority
vote per-repetition.

No data from the held-out subject is ever seen by our training code or model selection code at
any calibration budget; exactly the same windowing and evaluation criteria are applied equally
to our model as applied to all baseline methods.

Primary metrics reported for each method are macro-average F1-score (macro-$F_1$) with
associated distributions across subjects. Macro-$F_1$ is selected instead of pure accuracy due
to imbalance between gesture classes as well as between support counts per-gesture class; hence,
accuracy would misrepresent models favoring dominant classes.

\section{Experimental setup}
\label{sec:ch3-setup}

\paragraph{Datasets.} Four public datasets capture the variety of montages the encoder can
handle (Table~\ref{tbl:ch3-setup}(a)). The NinaPro
databases~\citep{atzori2014electromyography,pizzolato2017comparison} offer ten-, twelve-, and
sixteen-electrode arrangements recorded at sampling rates from 100~Hz to 2~kHz, while the
EMG-EPN612 database provides an eight channel armband containing recordings from a large
population of users. As such, these datasets represent a sampling space of different device
types, channel counts, and sampling frequencies that the encoder must support. Signals are
processed to ensure they remain in the Nyquist-safe frequency range and are resampled to a
uniform sampling frequency. Segments of signals are labeled using refinement-based repetition
labeling methods. Finally, segments of signals are windowed to a constant length.

\begin{table}[tp]\centering
\caption[Experimental setup]{Experimental setup. (a) Datasets used in this chapter. The spread of electrode counts and sampling rates is what the montage-agnostic encoder is required to span.
(b) Encoder architecture and the two training budgets used in this chapter. The scaled budget produces the headline cross-user results (Table~\ref{tbl:ch3-results}); the reduced budget was used for the component ablations and the self-supervised study, which are internally matched comparisons and are not compared numerically against the headline runs.}
\label{tbl:ch3-setup}

\begin{subtable}{\textwidth}\centering
\caption{}\label{tbl:ch3-setup-datasets}
\begin{tabular}{@{}lrrrr@{}}
\toprule
Dataset & Subjects & Electrodes & Rate (Hz) & Classes \\
\midrule
NinaPro DB1 & 27 & 10 & 100 & 53 \\
NinaPro DB2 & 40 & 12 & 2000 & 50 \\
NinaPro DB5 & 10 & 16 & 200 & 53 \\
EMG-EPN612 & 612 & 8 & 200 & 6 \\
\bottomrule
\end{tabular}
\end{subtable}

\vspace{1.5ex}

\begin{subtable}{\textwidth}\centering
\caption{}\label{tbl:ch3-setup-hyperparams}
{\footnotesize\setlength{\tabcolsep}{4pt}%
\begin{tabular}{@{}lcc@{}}
\toprule
Setting & Scaled budget & Reduced budget \\
\midrule
Encoder width $d$ & \multicolumn{2}{c}{256} \\
Transformer layers & \multicolumn{2}{c}{4} \\
Attention heads & \multicolumn{2}{c}{4} \\
Parameters & \multicolumn{2}{c}{3.3\,M} \\
Window / stride & \multicolumn{2}{c}{300\,ms / 150\,ms} \\
Common sample rate & \multicolumn{2}{c}{1\,kHz} \\
Optimiser & \multicolumn{2}{c}{AdamW ($\beta_1{=}0.9,\beta_2{=}0.98$)} \\
Schedule & \multicolumn{2}{c}{one-cycle} \\
Channel-masking rate & \multicolumn{2}{c}{0.25 (bounded, keep $\ge 3$)} \\
Label smoothing & \multicolumn{2}{c}{0.1} \\
Weight decay & \multicolumn{2}{c}{0.01} \\
\midrule
Base learning rate & $4\times10^{-4}$ & $4\times10^{-4}$ \\
Calibration learning rate & $3\times10^{-4}$ & $1\times10^{-4}$ \\
Base / calibration epochs & 50 / 30 & 25 / 15 \\
Training windows & 200\,000 & 60\,000 \\
Used for & headline cross-user runs & ablations, self-supervised study \\
\bottomrule
\end{tabular}}
\end{subtable}
\end{table}

\paragraph{Baseline.} All comparisons are relative to a per-user linear discriminant classifier
fit exclusively to the held-out user's own calibration repetitions using the Hudgins feature
set. This represents the real world clinic configured linear discriminants based solely upon the
user's calibration repetitions. This form of the LDA classifier was also measured to perform
better than when trained on a combined calibration window across multiple users. Therefore, we
report the per-user target-only LDA classifier as our baseline classifier.

\paragraph{Implementation.} The encoder has $d{=}256$, four transformer layers, and
approximately 3.3M total parameters. Training used the AdamW
optimizer~\citep{loshchilov2019decoupled} along with a one cycle schedule. All hyperparameter
settings are provided in Table~\ref{tbl:ch3-setup}(b). We ran experiments on a single local
GPU and the full leave-one-subject-out evaluations on A100 accelerators; the results matched.

\section{Results}

\subsection{Cross-user classification performance}
\label{sec:ch3-results-main}

\begin{figure}[tp]\centering
\begin{subfigure}{\textwidth}\centering
\includegraphics[width=0.98\textwidth]{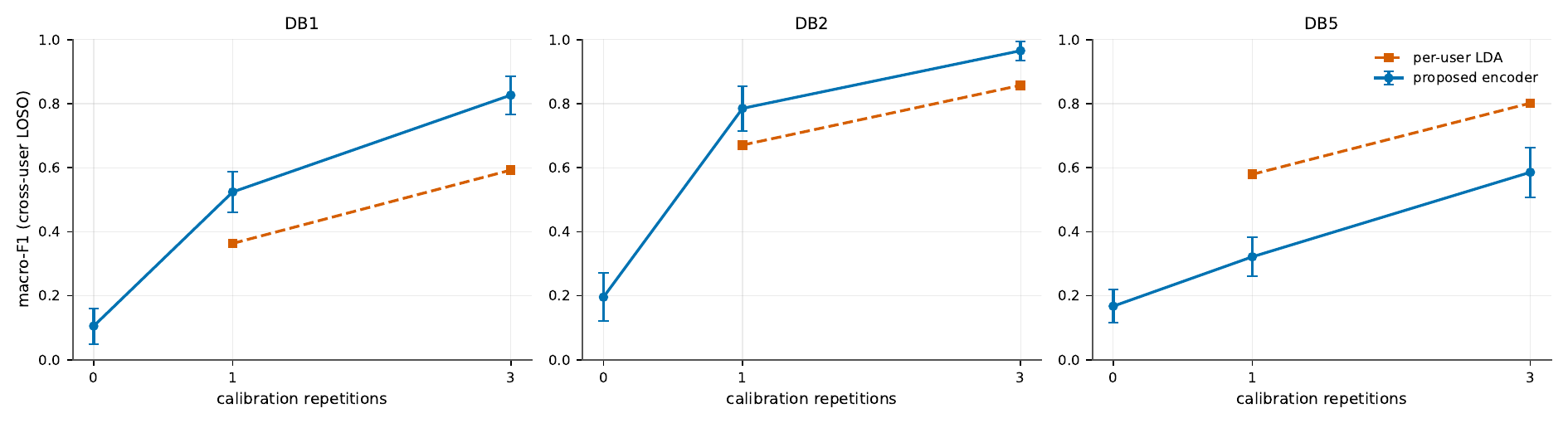}
\caption{}\label{fig:ch3-crossuser-calibration}
\end{subfigure}

\vspace{1.5ex}

\begin{subfigure}{\textwidth}\centering
\includegraphics[width=\textwidth]{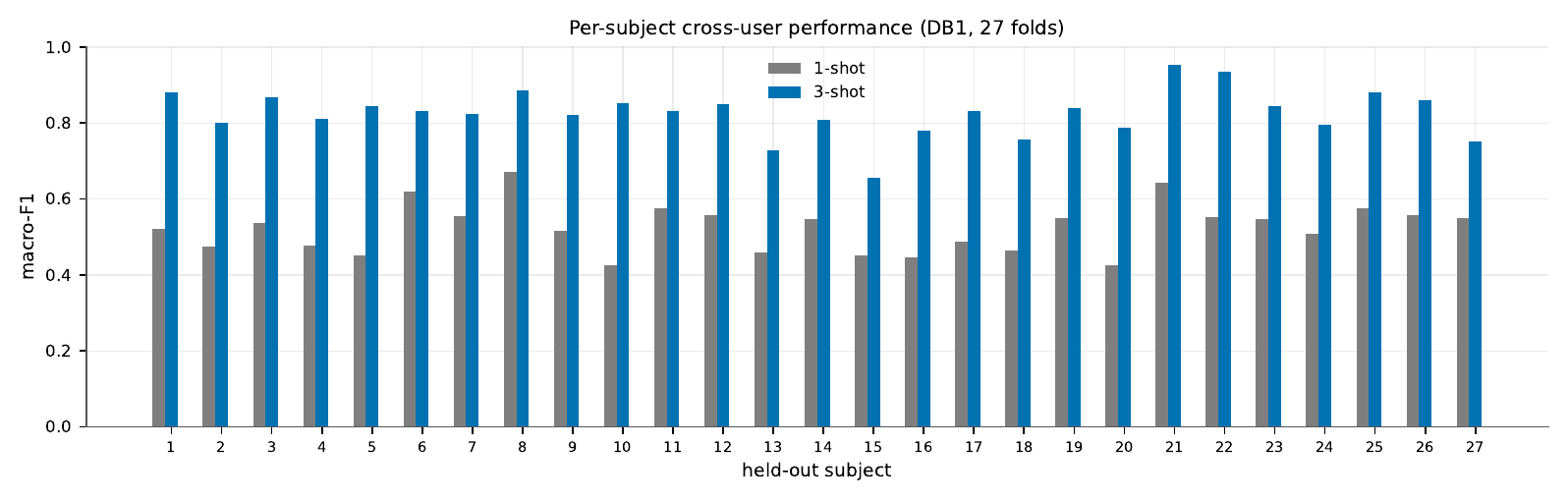}
\caption{}\label{fig:ch3-crossuser-persubject}
\end{subfigure}
\caption[Cross-user performance]{Cross-user performance. (a) Calibration-efficiency curves. Cross-user macro-F1 versus calibration budget for the proposed encoder and the per-user linear classifier, on each dataset. Error bars are the across-subject standard deviation.
(b) Per-subject cross-user macro-F1 on DB1. Every held-out subject exceeds the per-user LDA baseline at three calibration repetitions.}
\label{fig:ch3-crossuser}
\end{figure}

In Table~\ref{tbl:ch3-results} and Figure~\ref{fig:ch3-crossuser}(a), we provide the cross-user
classification results using the full 27-subject NinaPro DB1 dataset. We first consider the
cross-user performance at three calibration repetitions. The proposed encoder achieves a
macro-F1 value of $0.827 \pm 0.060$ at three repetitions, significantly outperforming the
$0.593$ of the per-user LDA classifier fitted and evaluated on the same 27 held-out users, a
margin of $+0.234$. When evaluating the cross-user performance at a single repetition, the
encoder achieves a macro-F1 value of $0.524$ vs $0.363$ for the per-user LDA classifier (i.e.,
an improvement of $+0.161$). Notably, the zero-calibration F1 score is $0.105$, far exceeding a
chance level for recognizing among 52 fine-grained hand gestures in a completely unknown user
without any calibration.

\begin{table}[tp]\centering
\caption[Cross-user results]{Cross-user leave-one-subject-out macro-F1 by calibration budget. The comparison baseline is a per-user LDA trained on the held-out user's own calibration repetitions, fitted and scored over the same held-out subjects as the encoder. Right-hand block: paired comparison of the encoder against the per-user linear classifier at three calibration repetitions, matched by held-out subject. $p$ from a two-sided Wilcoxon signed-rank test.}
\label{tbl:ch3-results}
\adjustbox{max width=\textwidth}{%
\begin{tabular}{@{}llcccrccc@{}}
\toprule
 & & \multicolumn{3}{c}{macro-F1 by calibration budget}
   & \multicolumn{4}{c}{paired test at 3 repetitions} \\
\cmidrule(lr){3-5}\cmidrule(lr){6-9}
Dataset & Model & 0-shot & 1-shot & 3-shot & $n$ & Median $\Delta$ & Subjects ahead & $p$ \\
\midrule
DB1 & proposed & $0.105\pm0.056$ & $0.524\pm0.063$ & $0.827\pm0.060$
  & \multirow{2}{*}{27} & \multirow{2}{*}{$+0.219$} & \multirow{2}{*}{27/27} & \multirow{2}{*}{$5.6e-06$} \\
 & per-user LDA & -- & $0.363$ & $0.593$ & & & & \\
\addlinespace
DB2 & proposed & $0.196\pm0.075$ & $0.785\pm0.070$ & $0.965\pm0.030$
  & \multirow{2}{*}{40} & \multirow{2}{*}{$+0.100$} & \multirow{2}{*}{39/40} & \multirow{2}{*}{$4.8e-08$} \\
 & per-user LDA & -- & $0.671$ & $0.857$ & & & & \\
\addlinespace
DB5 & proposed & $0.168\pm0.051$ & $0.322\pm0.061$ & $0.586\pm0.077$
  & \multirow{2}{*}{10} & \multirow{2}{*}{$-0.208$} & \multirow{2}{*}{0/10} & \multirow{2}{*}{$2.0e-03$} \\
 & per-user LDA & -- & $0.579$ & $0.802$ & & & & \\
\bottomrule
\end{tabular}}
\end{table}

As presented in Table~\ref{tbl:ch3-persubject} and Fig.~\ref{fig:ch3-crossuser}(b), all 27
held-out users achieve higher macro-F1 values than the per-user LDA classifier at three
repetitions (ranging from $0.656$ to $0.953$).

\begin{table}[t]\centering\small
\caption[Per-subject results]{Per-subject cross-user results on DB1. Every held-out subject is reported so the spread behind the mean is visible. Each subject contributes 156 trials.}
\label{tbl:ch3-persubject}
\begin{tabular}{@{}rcccc@{}}
\toprule
Subject & 0-shot & 1-shot & 3-shot & bal.\ acc.\ (3-shot) \\
\midrule
1 & 0.081 & 0.523 & 0.882 & 0.885 \\
2 & 0.023 & 0.474 & 0.800 & 0.814 \\
3 & 0.058 & 0.536 & 0.868 & 0.872 \\
4 & 0.082 & 0.479 & 0.811 & 0.821 \\
5 & 0.080 & 0.452 & 0.844 & 0.853 \\
6 & 0.106 & 0.619 & 0.833 & 0.846 \\
7 & 0.048 & 0.555 & 0.824 & 0.840 \\
8 & 0.152 & 0.671 & 0.885 & 0.891 \\
9 & 0.057 & 0.517 & 0.821 & 0.833 \\
10 & 0.112 & 0.425 & 0.853 & 0.859 \\
11 & 0.084 & 0.577 & 0.833 & 0.846 \\
12 & 0.116 & 0.558 & 0.849 & 0.853 \\
13 & 0.055 & 0.459 & 0.729 & 0.750 \\
14 & 0.105 & 0.547 & 0.809 & 0.833 \\
15 & 0.097 & 0.451 & 0.656 & 0.686 \\
16 & 0.117 & 0.447 & 0.781 & 0.801 \\
17 & 0.255 & 0.488 & 0.832 & 0.846 \\
18 & 0.218 & 0.465 & 0.757 & 0.776 \\
19 & 0.116 & 0.551 & 0.839 & 0.846 \\
20 & 0.074 & 0.425 & 0.789 & 0.801 \\
21 & 0.184 & 0.642 & 0.953 & 0.955 \\
22 & 0.044 & 0.553 & 0.935 & 0.936 \\
23 & 0.139 & 0.548 & 0.846 & 0.853 \\
24 & 0.079 & 0.508 & 0.796 & 0.801 \\
25 & 0.186 & 0.575 & 0.881 & 0.891 \\
26 & 0.149 & 0.559 & 0.859 & 0.865 \\
27 & 0.033 & 0.550 & 0.752 & 0.756 \\
\midrule
mean & 0.105 & 0.524 & 0.827 & \\
s.d. & 0.056 & 0.063 & 0.060 & \\
\bottomrule
\end{tabular}
\end{table}

We further perform a two-sided Wilcoxon signed-rank test on the per-subject difference between
the encoder's macro-F1 and that of the per-user LDA classifier at three repetitions. The test
rejects equality for those datasets where the encoder is superior (see
Table~\ref{tbl:ch3-results}).

\subsection{Component ablations}
\label{sec:ch3-ablation}

To assess the importance of each component of the encoder, we sequentially remove each of the
three core components at the reduced training budget described in
Table~\ref{tbl:ch3-setup}(b), resulting in versions that are comparable to one another. Each
variant replaces rolling-time normalization with a simple z-scoring operation applied to each
time window individually, removes the spatial position representation of electrodes, or
eliminates cross-channel attention (see Table~\ref{tbl:ch3-ablation} and
Fig.~\ref{fig:ch3-training}(a)) -- each version loses more than half of the encoder's three-shot
macro-F1 value. These ablated versions do not collapse to a single class; instead, they retain
enough ability to classify many classes well above chance levels, implying that these losses
represent losses of capacity rather than failures to train. Importantly, cross-channel attention
is the single largest contributor to this lost capacity, followed closely by rolling-time
normalization and finally by spatial information about electrode positions.

Importantly, if we sum up the individual losses, they contribute to more than the entire model's
score, indicating that these components are complementary rather than independent contributors
to overall performance.

\begin{table}[t]\centering
\caption[Component ablations]{Component ablations on DB1 (cross-user LOSO). $\Delta$ is the change from the full model at each calibration budget.}
\label{tbl:ch3-ablation}
{\footnotesize\setlength{\tabcolsep}{4pt}%
\begin{tabular}{@{}lcccccc@{}}
\toprule
Variant & 0-shot & $\Delta$ & 1-shot & $\Delta$ & 3-shot & $\Delta$ \\
\midrule
full model & 0.081 & -- & 0.331 & -- & 0.677 & -- \\
without rolling-time normalisation & 0.033 & $-0.048$ & 0.042 & $-0.289$ & 0.113 & $-0.564$ \\
without electrode-coordinate encoding & 0.027 & $-0.055$ & 0.066 & $-0.264$ & 0.142 & $-0.535$ \\
without cross-channel attention & 0.030 & $-0.051$ & 0.032 & $-0.299$ & 0.063 & $-0.614$ \\
\bottomrule
\end{tabular}}
\end{table}

\subsection{Effect of training budget}
\label{sec:ch3-scaling}

\begin{figure}[tp]\centering
\begin{subfigure}{\textwidth}\centering
\includegraphics[width=0.86\textwidth]{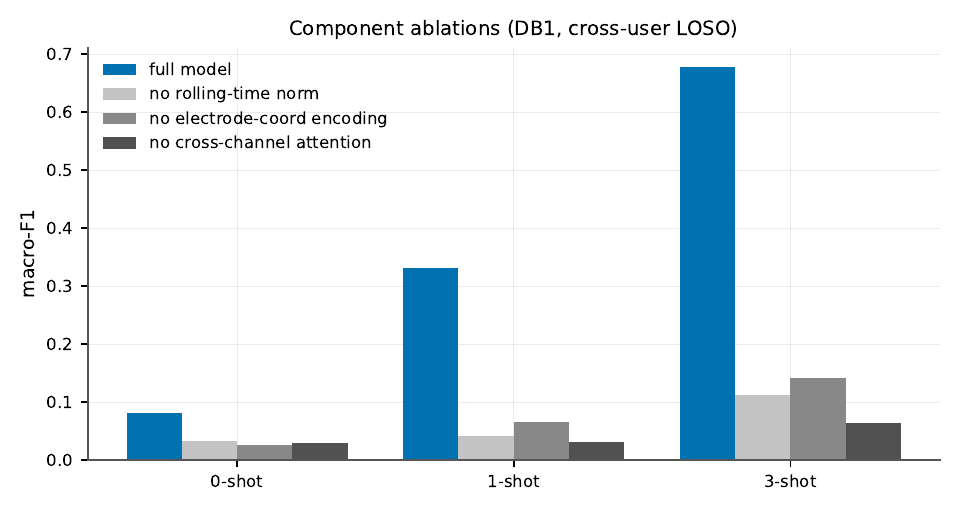}
\caption{}\label{fig:ch3-training-ablation}
\end{subfigure}

\vspace{1.5ex}

\begin{subfigure}{0.49\textwidth}\centering
\includegraphics[width=\textwidth]{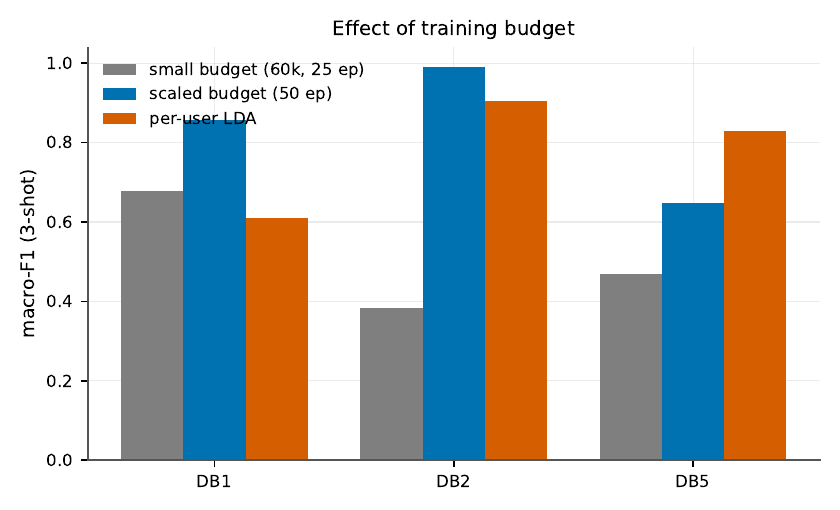}
\caption{}\label{fig:ch3-training-scaling}
\end{subfigure}\hfill
\begin{subfigure}{0.49\textwidth}\centering
\includegraphics[width=\textwidth]{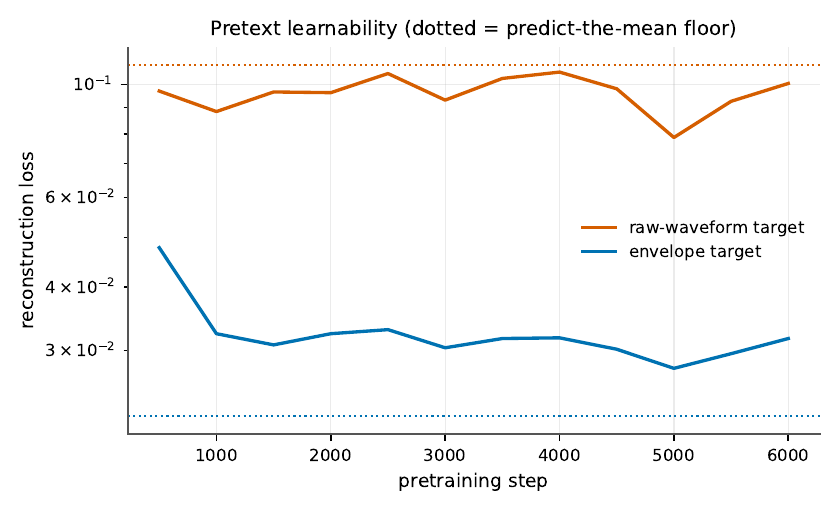}
\caption{}\label{fig:ch3-training-pretext}
\end{subfigure}
\caption[Training-side studies]{Training-side studies. (a) Component ablations on DB1. Removing rolling-time normalisation, the electrode-coordinate position encoding, or cross-channel attention each costs more than half of the model's three-shot macro-F1.
(b) Effect of training budget on the comparison against the LDA baseline. Under-training reverses the outcome on the higher-fidelity datasets.
(c) Pretext learnability. Raw-waveform reconstruction sits at its predict-the-mean floor and learns nothing; the envelope objective drops well below its floor and learns.}
\label{fig:ch3-training}
\end{figure}

The comparison against the LDA baseline is not fixed; it is determined by how much training is
performed (Table~\ref{tbl:ch3-scaling} and Fig.~\ref{fig:ch3-training}(b)). On DB2, the
highest-resolution dataset, an encoder trained with limited resources falls $-0.52$ below a
per-user LDA classifier at three repetitions. However, when trained sufficiently on DB2 it
surpasses the per-user LDA classifier and its score exceeds $0.98$ at three repetitions. The
direction of the comparison therefore reverses with training budget: what appears to be a large
margin favoring the LDA classifier is merely due to inadequate training of the trainable model.
This effect occurs again in Section~\ref{sec:ch3-pretrain}, in which it accounts for an observed
low-calibration gap previously misattributed to an omitted pretraining prior.

\begin{table}[t]\centering
\caption[Training budget]{Effect of training budget on the comparison against the LDA baseline. Under-training reverses the outcome on the higher-fidelity datasets.}
\label{tbl:ch3-scaling}
\begin{tabular}{@{}lcccc@{}}
\toprule
Dataset & Small budget & Scaled budget & per-user LDA & $\Delta$ (scaled) \\
\midrule
DB1 & 0.677 & 0.857 & 0.610 & $+0.247$ \\
DB2 & 0.383 & 0.989 & 0.904 & $+0.085$ \\
DB5 & 0.468 & 0.647 & 0.830 & $-0.182$ \\
\bottomrule
\end{tabular}
\end{table}

\subsection{Subject count and baseline strength}
\label{sec:ch3-interaction}

There is no single answer among the three datasets studied. The encoder outperforms the per-user
LDA baseline substantially on DB1, slightly on DB2 after being trained adequately, and falls
behind it on DB5. Notably, none of these outcomes depend on sampling rate or number of channels
directly and not even on number of training subjects over the range available, as demonstrated
in the subject-sweep experiment below. What tracks that ordering is the strength of the per-user
LDA baseline.

\begin{figure}[tp]\centering
\begin{subfigure}{\textwidth}\centering
\includegraphics[width=0.76\textwidth]{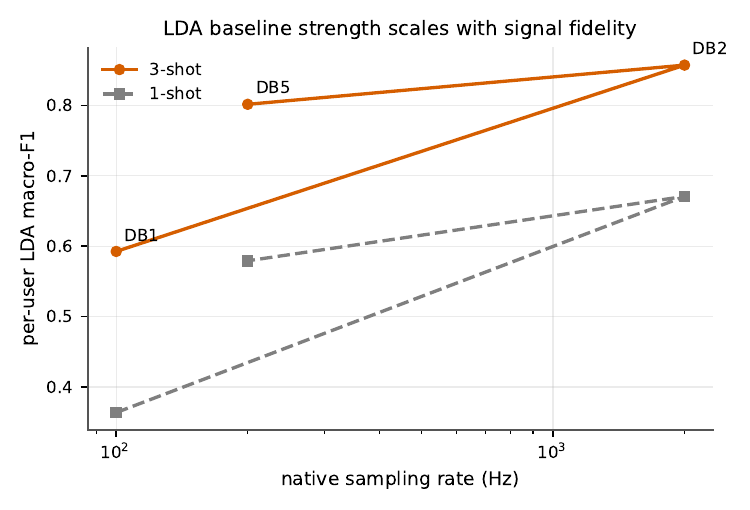}
\caption{}\label{fig:ch3-margin-fidelity}
\end{subfigure}

\vspace{1.5ex}

\begin{subfigure}{\textwidth}\centering
\includegraphics[width=0.76\textwidth]{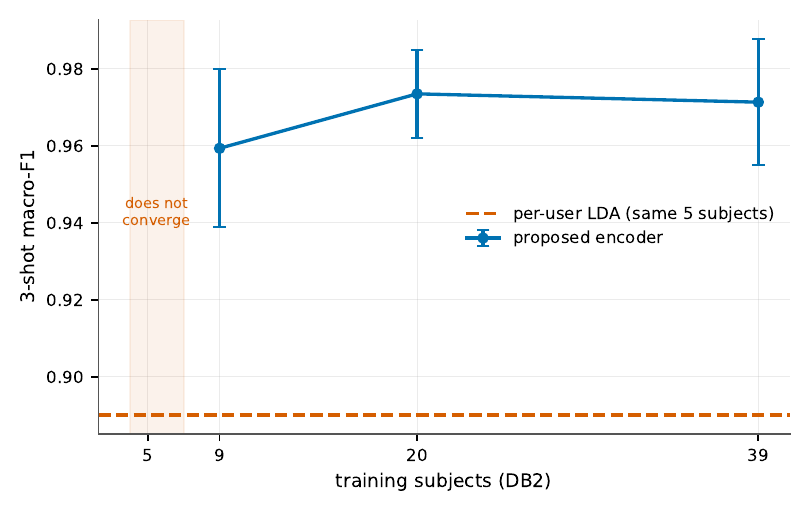}
\caption{}\label{fig:ch3-margin-diversity}
\end{subfigure}
\caption[Baseline strength and training-pool size]{Baseline strength and training-pool size.
(a) The LDA baseline strengthens with signal fidelity: per-user LDA macro-F1 rises with sampling rate across DB1 (100 Hz), DB5 (200 Hz), and DB2 (2 kHz).
(b) DB2 subject-count sweep at a convergent learning rate. The margin over the per-user baseline is close to flat from nine training subjects upward; at five the model does not converge at all, so the constraint is a stability floor rather than a smooth decline.}
\label{fig:ch3-margin}
\end{figure}

\begin{figure}[t]\centering
\includegraphics[width=0.98\textwidth]{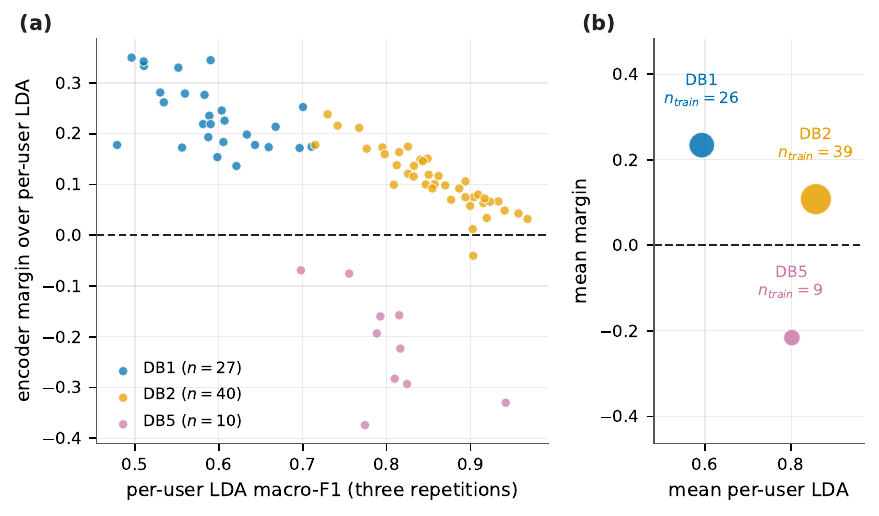}
\caption{Encoder margin against the strength of the per-user baseline, on the intact databases. (a) One point per held-out subject: the per-user LDA score at three calibration repetitions against the encoder's margin over it, for all 77 held-out subjects. Points below the dashed line are subjects where the per-user classifier is ahead. (b) Database means, with marker area proportional to the leave-one-subject-out training pool. DB2 has a stronger baseline than DB5 yet a positive margin, and DB5 is also the database at the subject-count floor of Section~\ref{sec:ch3-interaction}.}\label{fig:ch3-baseline-margin}
\end{figure}

Fig.~\ref{fig:ch3-margin}(a) demonstrates that as signal quality improves (measured through
fidelity), so too does performance of the per-user LDA baseline: on DB1, it achieves an average
macro-F1 value of $0.593$ at 100~Hz; on DB5, it achieves an average macro-F1 value of $0.802$ at
200~Hz; and on DB2 it achieves an average macro-F1 value of $0.857$ at 2~kHz. Hand-crafted
features obviously have greater bandwidth to capitalize upon as sampling rates increase. Thus,
the margins open to trainable encoders are least when baselines are strongest; see
Fig.~\ref{fig:ch3-baseline-margin} for a breakdown by subject.


DB2 is used for testing the second part of this explanation regarding subject diversity; see
Table~\ref{tbl:ch3-diversity} and Fig.~\ref{fig:ch3-margin}(b). Holding out the exact same five
subjects across all iterations and comparing to a per-user LDA classifier fit on those same five
subjects yields a three-shot margin for the encoder of $+0.081$ with thirty-nine training
subjects, $+0.083$ with twenty training subjects, and $+0.069$ with nine training subjects.
Within this range, the gain remains nearly flat: cutting off roughly seventy-five percent of
training subjects reduces margin by about a point and a half. With only five training subjects
however, the model fails to converge at all (i.e., it produces a single-class predictor on all
five folds at both tested learning rates).

Therefore, constraints placed by limiting subject diversity are a stability bound rather than a
continuous decline. Down to nine subjects however, the encoder retains most of its advantages
versus the per-user classifier; below that it ceases training altogether as opposed to degrading
progressively. An earlier iteration of this subject-sweep experiment, conducted at a learning
rate of $4\times10^{-4}$, appeared to show a sharp dose response curve including a large loss at
nine subjects. This apparent effect arose from running training sessions that collapsed down to
a single class predictor that were averaged together with converged sessions; at
$1.5\times10^{-4}$ however, the same condition converges on every fold while remaining ahead of
the baseline. The lower learning rate is necessary because reducing subject count reduces
training-set size, and small amounts of data cannot withstand schedules designed for larger
sets.

Thus, across all three datasets, what continues to hold true is the baseline-strength half of
this account. The per-user classifier is weaker on DB1 (F1 = $0.593$) compared to DB5 (F1 =
$0.802$) and DB2 (F1 = $0.857$); this ordering among datasets corresponds to the experimental
outcome: substantial margin on DB1; lesser margin on DB2; and a loss on DB5 with only ten
subjects available (i.e., operating at the edge of conditions required for stable training).
This is an observational contrast rather than a controlled one, since sampling rate, class count
and subject pool all differ between databases.

\begin{table}[t]\centering
\caption[Subject-count sweep]{DB2 subject-count sweep, five held-out subjects per condition, scored against a per-user LDA baseline of 0.890 fitted on the same five. The encoder's margin is close to flat from nine training subjects upward; below that, cross-user training does not converge at this model size, at either learning rate tried.}
\label{tbl:ch3-diversity}
\begin{tabular}{@{}rccl@{}}
\toprule
Train subjects & 3-shot & $\Delta$ & Note \\
\midrule
39 & 0.971 & $+0.081$ & -- \\
20 & 0.973 & $+0.083$ & -- \\
9 & 0.959 & $+0.069$ & -- \\
5 & -- & -- & did not converge (5/5 folds) \\
\bottomrule
\end{tabular}
\end{table}

\subsection{Self-supervised pretraining}
\label{sec:ch3-pretrain}

The other direction to strengthen the low-calibration side is to train the encoder using
unlabelled pooled EMG data via self-supervised pretraining. I attempted two tasks, but neither
improved downstream accuracy (Table~\ref{tbl:ch3-pretrain}, Figure~\ref{fig:ch3-training}(c)).

The first task was masked reconstruction of the original waveform. However, this task is
degenerate. The loss function for reconstructing the per-channel mean of the target is $0.109$.
The pretext task converged exactly to this value and remained at this level: since sample-level
EMG can be described as nearly random, the reconstruction provides no useful information to the
encoder. When changing the object of the loss function to the signal envelope (root-mean-squared
power over short sub-windows), I obtained some form of learning. Specifically, I achieved a
seventeen-fold decrease in the loss function toward a floor value determined based on oracle
knowledge that the model has never seen. Therefore, the signal envelope is a possible
self-supervised target for learning on this signal where the raw waveform is not.

Although the second task learned something, it didn't transfer. Training the encoder with
weights initialized from those previously trained using the envelope-loss resulted in no change
in downstream cross-user accuracy within noise. Constraining the encoder to remain near these
initial weights during fine-tuning made things even worse, ruling out the possibility that the
fine-tuning schedule simply washed out the prior knowledge and instead pointing towards a lack
of correspondence between the prior and the cross-user task. The scaling result above explains
why: the low-calibration gap that motivated this pretraining was artificial and disappeared once
the supervised model was sufficiently well-trained. At this scale, with this signal, a
supervised montage-aware encoder trained on enough subjects doesn't require any kind of
self-supervised prior.

\begin{table}[t]\centering
\caption[Pretraining objectives]{Self-supervised pretraining objectives evaluated on DB1. The reconstruction floor is the loss obtained by predicting the mean of the target; a pretext that converges to its floor has learned nothing.}
\label{tbl:ch3-pretrain}
\begin{tabular}{@{}lccccc@{}}
\toprule
Pretext target & Final loss & Floor & Learns? & 1-shot & 3-shot \\
\midrule
none (from scratch) & -- & -- & -- & 0.331 & 0.677 \\
raw waveform & 0.100 & 0.109 & no & 0.344 & 0.628 \\
envelope (RMS) & 0.030 & 0.022 & yes & 0.309 & 0.668 \\
\bottomrule
\end{tabular}
\end{table}

\section{Discussion}
\label{sec:ch3-discussion}

A single montage-aware encoder that adapts to a new user after a few repetitions from that user
surpasses a per-user linear classifier in terms of cross-user gesture recognition accuracy when
the encoder is fully trained. Although the three databases tested have different strengths for
their respective per-user baselines (which depend upon how much fidelity each signal contains),
it is that strength, not the number of training subjects, that tracks the outcome. In
particular, since I conducted a controlled experiment to test whether or not subject count
impacts performance (i.e., I tested my system on DB2 while keeping the per-user baseline
constant at $0.890$, and found that the 3-shot margin for the encoder was $+0.069$ at 9 training
subjects, $+0.083$ at 20 training subjects and $+0.081$ at 39 training subjects), I know that
subject count impacts performance only as a floor effect. That is, if training occurs with fewer
than about 10 subjects, then training fails completely (i.e., it converges to a one-class
predictor), rather than failing gracefully. Therefore, an encoder of this size does not require
40 subjects; approximately 10 do suffice, however they must be trained with an optimizer retuned
for lower-pool sizes. More broadly, a study that included only DB1 would report a clear and
universal victory for a decoder-based approach; whereas a study that only included DB5 would
report that per-user LDA calibration remains unbeatable. Neither study would be correct. The
studies' disagreements would be resolved based on how strong the per-user baseline is for each
study, and specifically DB5 would sit at the boundary of the subject-count floor rather than
being influenced solely by either individual study's results.

Three limitations restrict the implications of the results presented in this paper. First, the
evaluation presented here is offline; i.e., it presents decoding accuracy under a specific
protocol, not usability metrics associated with wearing a device in real-time. Second,
EMG-EPN612 gave an ambiguous result, which appeared low for both the encoder and the LDA
baseline. This ambiguity arose because this study uses windowing across entire recordings rather
than segmenting gestures into discrete units. Since the latter method is more commonly employed,
this study reports results in order to provide additional context; they should not be compared
to previous results based on this same dataset's protocol. Third, the spatial positions
represented by electrodes in this study are approximate.

\section{Conclusion}
\label{sec:ch3-conclusion}

In this paper, I developed a montage-agnostic encoder for calibration-light cross-user gesture
recognition and demonstrated that this encoder exceeds a per-user linear classifier with respect
to cross-user gesture recognition on all users of the full NinaPro DB1 benchmark and using just
three calibration repetitions. Each of the encoder's three key components individually accounts
for more than half of its 3-shot macro F1 in an otherwise budget-matched ablation study. I also
demonstrated that comparing against an LDA baseline depends on budget spent training models and
on how good that baseline is, and that self-supervised pretraining had no benefits once a
supervised model was adequately trained.

\section*{CRediT authorship contribution statement}
\textbf{Jethro Odeyemi:} Conceptualization, Methodology, Software, Validation, Formal analysis,
Investigation, Data curation, Writing -- original draft, Writing -- review and editing,
Visualization.

\textbf{W.J. (Chris) Zhang:} Conceptualization, Methodology, Resources, Supervision,
Project administration, Writing -- review and editing.

\section*{Declaration of competing interest}
The authors declare no competing financial interests or personal relationships that could have
appeared to influence the work reported in this paper.

\section*{Funding}
This research did not receive any specific grant from funding agencies in the public, commercial, or not-for-profit sectors.

\section*{Ethics statement}
This work involved no new data collection from human participants. All analyses were performed
on previously published, publicly released surface-EMG databases, for which ethical approval and
informed consent were obtained by the original data collectors.

\section*{Data availability}
This study uses only publicly available data. The NinaPro databases are available from the
NinaPro consortium and the EMG-EPN612 database from its original distributors; the specific
releases used are cited in Section~\ref{sec:ch3-setup}.
No new data were generated by this study.

\bibliographystyle{elsarticle-num-names}
\bibliography{refs}

\end{document}